\documentclass[letterpaper]{article} % DO NOT CHANGE THIS
\usepackage{aaai24}  % DO NOT CHANGE THIS
\usepackage{times}  % DO NOT CHANGE THIS
\usepackage{helvet}  % DO NOT CHANGE THIS
\usepackage{courier}  % DO NOT CHANGE THIS
\usepackage[hyphens]{url}  % DO NOT CHANGE THIS
\usepackage{graphicx} % DO NOT CHANGE THIS
\usepackage{natbib}  % DO NOT CHANGE THIS AND DO NOT ADD ANY OPTIONS TO IT
\usepackage{caption} % DO NOT CHANGE THIS AND DO NOT ADD ANY OPTIONS TO IT
\usepackage{algorithm}
\usepackage{algorithmic}

\usepackage{newfloat}
\usepackage{listings}
\DeclareCaptionStyle{ruled}{labelfont=normalfont,labelsep=colon,strut=off} % DO NOT CHANGE THIS
\floatstyle{ruled}
\newfloat{listing}{tb}{lst}{}
\floatname{listing}{Listing}
\usepackage[usestackEOL]{stackengine}
\usepackage{bm}
\usepackage{eqnarray}
\usepackage{amsmath,amssymb,amsfonts}
\usepackage{mathtools}
\usepackage{xcolor}

\title{A Framework for Exploring Federated Community Detection}
\author{
    William Leeney,
    Ryan McConville
}
\affiliations{
    University of Bristol\\
    will.leeney@bristol.ac.uk, ryan.mcconville@bristol.ac.uk
}

\begin{document}

\maketitle

\begin{abstract}

Federated Learning is machine learning in the context of a network of clients whilst maintaining data residency and/or privacy constraints. Community detection is the unsupervised discovery of clusters of nodes within graph-structured data. The intersection of these two fields uncovers much opportunity, but also challenge. For example, it adds complexity due to missing connectivity information between privately held graphs. In this work, we explore the potential of federated community detection by conducting initial experiments across a range of existing datasets that showcase the gap in performance introduced by the distributed data. We demonstrate that isolated models would benefit from collaboration establishing a framework for investigating challenges within this domain. The intricacies of these research frontiers are discussed alongside proposed solutions to these issues.  

\end{abstract}

\section{Introduction}

Data ownership can be distributed across institutions or individuals who may not be able to share it for technical or legal reasons to enable the use of centralised machine learning \cite{liu2022federated,tan2022towards}. Federated learning over graphs is complex due to the variety of scenarios created by partitions of this structure  \cite{zhang2021federated}. GNNs are used extensively in previous literature for learning network representations \cite{chen2020graph,hamilton2017representation,scarselli2008graph,kipf2016semi} and there are many works studying the unsupervised use of these methods \cite{tsitsulin2020graph,wang2019attributed}. Identifying clusters based on latent patterns in the dataset, rather than the ``ground-truth" is useful for applications with missing or prohibited information access. This makes community detection a powerful tool for studying federated learning on graphs. Numerous surveys exist on federated learning methods \cite{ma2022state,yang2019federated} and graph federated learning \cite{yang2019federated,liu2022federated} but none on federated community detection. 

Federated graph learning encompasses a diverse range of scenarios, each with important applications. For banks that want to collaborate for anti-fraud measures but can't share client information \cite{suzumura2019towards} it is useful to study the setup where each client owns a sub-section of a global graph. Social media networks suffer from fake news propagating between the networks \cite{han2020graph,liu2022federatedsocial} but data is split heterogeneous between clients as a comment on one platform is not the same as a retweet. Pharmaceutical companies own molecular copyright and don't want to share \cite{manu2021fl} meaning that they are aware of the space graph structure but do not want to share important feature space information that is expensive to gather. Hospitals cannot share sensitive information but working together would improve patient care so studying a partition where there are missing connections between nodes is helpful here. This federated paradigm is also applicable to learning a personalised model where each client is a user of the same platform. It is desirable to learn a model specific to each user whilst also sharing global information insights which is pertinent in context of data privacy regulations such as the General Data Protection Regulation (GDPR) \cite{truong2021privacy}. Therefore, this paradigm also can extend to learning topology of clients to facilitate better communication between a network of clients.

We explore an initial scenario and method for unsupervised federated graph clustering which averages the weights of clients \cite{mcmahan2017communication} with a modularity based clustering objective \cite{tsitsulin2020graph}. We test the method against clients that do not collaborate over three datasets and demonstrate that the federation leads to better performance than the isolated case but still lacks compared to the centralised alternative. This work creates a benchmark and framework on which other graph clustering objectives and federation approaches can be used. Solutions have been proposed that use GNNs for supervised federated graph learning solving various sub-problems that can be adapted for the unsupervised version of the problem such as: personalisation \cite{xie2021federated}, non-identically distribution graphs \cite{caldarola2021cluster}, multi-task and multi-domain federation learning \cite{zhuang2021towards}. Some unsupervised federated techniques that have already been applied to images include non-IID data \cite{zhang2020federated} and for unbalanced data \cite{servetnyk2020unsupervised}. 

We will discuss open challenges directly related to graph federated learning that become increasingly difficult without explicit labels such as malicious attack vulnerability and distributed communication protocols in combination with solutions to alleviate these.

\section{A Methodology For Experimentation}

\subsection{Baseline Problem Statement}

We consider the simplest version of federated graph learning which is that the problem is defined by having number of clients $C$, which each own sub-graphs $\mathcal{G} = [\mathcal{G}^1, \mathcal{G}^2,... \mathcal{G}^C] = [\{\bm{X}^1, \bm{A}^1\}, \{\bm{X}^2, \bm{A}^2\}, ..., \{\bm{X}^C, \bm{A}^C\}]$ of a global graph. Each graph $\mathcal{G}^c = (\mathcal{V}^{c}, \mathcal{E}^c)$ is represented by a set of nodes $u^c \in \mathcal{V}^c$ and a set of edges $(u^c, v^c) \in \mathcal{E}^c$ that depends on the set of $|\mathcal{V}^c| = N^c$ nodes allocated to each client which also dictates the number of total $|\mathcal{E}^c| = M^c$ edges but all own the same set of $k$ features. This allows us to represent the each clients graph as $\mathcal{G}^c = \{ \bm{X}^{c}, \bm{A}^{c} \}$, where $\bm{X}^{c} \in \mathbb{R}^{N^c \times k}$ is the node feature matrix and the adjacency matrix is represented by $\bm{A}^{c} \in \mathbb{R}^{N^c \times N^c}$. A node $u_{i}^{c}$ has a degree $d_{u}^{c}=\textstyle\sum_{v=1}^{N^c}\mathbf{A}_{u v}^{c}$ which is the number of nodes it is connected to $|\mathcal{N}_{u}^{c}| = d_{u}^{c}$. The degree vector is therefore given by $\bm{D}^{c} = [ d_{1}^{c}, d_{2}^{c}, ..., d_{N^{c}}^{c}]^\top$. We seek to learn a function for each client $c$ to find cluster assignments $\bm{Y}^{c} \in \{1, 2, ..., k\}^{N}$ for the $k$ clusters present in the data. In practise, each client finds their own cluster assignment, so the size of the matrix will depend on the number of nodes each client $c$ has, denoted by $N^c$.

\subsection{Datasets}

In these experiments the models are tested over three different datasets, Cora \cite{mccallum2000automating}, Citeseer \cite{giles1998citeseer} and PubMed \cite{sen2008collective} which are split via nodes to create the dataset partitions for each client. This allows a direct comparison to the typical GNN clustering algorithm which these are usually used to test which mimics the centralised alternative. Each graph is a publication citation network with features extracted from the words in the papers. We contextualise these graphs in terms of potential graph topologies as the partitioning scheme may result in non-isomorphic graph compared to the original. The average clustering coefficient is the average of all the local clustering coefficients calculated for each node, which is the proportion of all the connections that exist in a nodes neighbourhood, over all the links that could exist in that neighbourhood. Closeness centrality is defined as the reciprocal of the mean shortest path distance from all other nodes in the graph, so the average of this is a measure of how close the average node is to other nodes.

\renewcommand{\arraystretch}{1.2}
\begin{table}[h]
    \centering
    \begin{tabular*}{\dimexpr\linewidth}{c|ccc}
    \hline
    \textbf{Datasets}  &   \textbf{Cora }   &   \textbf{CiteSeer}   &   \textbf{PubMed}  \\ 
    \hline
    \hline
    N Nodes   &   2708     &   3327     &      19717 \\ 
    N Features   &  1433      &    3703    &     500 \\ 
    N Edges   &   10556     &   9104     &    88648 \\ 
    \Longunderstack{Mean Closeness \\ Centrality}  & 0.137 & 0.047      &  0.160  \\ 
    \Longunderstack{Average Clustering \\ Coefficient}  &   0.241     &  0.144      &   0.060 \\ 
    \end{tabular*}
    \caption{ The datasets, and associated statistics, we will study.}
    \label{tab:stats}
\end{table}

\subsection{Unsupervised GNN Clustering Architecture}

Here the clustering objective and forward pass of each GNN that is optimised by each client is described and for simplicity we omit the client superscript. In these experiments, the same model is used for each client as the federated part of the algorithm averages weights. There is potential to use different GNN models if the federation aggregation used an embedded space minimisation \cite{makhija2022architecture}. 

The graph neural network (GNN) computes a function $\bm{F}(\bm{X}) = [\bm{h}_1 \bm{h}_2,...,\bm{h}_N ]^T $ over every node's neighbourhood $\boldsymbol{h_u} = \phi(\bm{X}_{\mathcal{N}_u})$. Computing the function over the features at layer $l$ gives the features at the next layer and the permutation-invariant aggregator operator that we use is $\sum$. The first step in the GNN architecture used is to transform the adjacency matrix $\bm{A}$ by adding self loops then normalising by the degree matrix so $\bm{\hat{A}} = \bm{D}^{-\frac{1}{2}}(\bm{A} + \bm{I})\bm{D}^{-\frac{1}{2}}$. The neighbourhood of a node $\mathcal{N}_u = \{v | \hat{a}_{vu} > 0 \}$ as the node's that it is connected to. The neighbourhood feature matrix for a node is the multi-set of all neighbourhood features $\bm{X}_{\mathcal{N}_u} = \{ \{ \bm{x}_v | v \in \mathcal{N}_u \} \}$, which includes the features of node $u$. We use a single layer GNN so the hidden representation for a single node $u$ is given by $\boldsymbol{h_u} = \sigma(b_u + \sum_{v \in \mathcal{N}_u} \hat{a}_{uv} \cdot \psi(\bm{x}_v))$, where $\psi(\bm{x}) = \bm{Wx}$; $b_u$ is the bias for the node and the chosen activation function $\sigma()$ is SELU \cite{klambauer2017self}. 

The forward pass is given by $\bm{Y} = \xi(\bm{F}(\bm{X}))$, where $\xi(\bm{x}) = U(\bm{W}\bm{x} + \bm{b})$ is a neural network and $U()$ is the Softmax activation function, which is chosen as to find cluster assignments $\bm{Y}$. The algorithm optimises modularity given by Equation \eqref{equ: modularity}, where the function $\delta(c_{u},c_{v}) = 1$ if nodes $u$ and $v$ belong the same cluster and $0$ otherwise. Modularity measures the divergence between the intra-cluster edges from the expected one under a random graph with given degrees $d_u$ and $d_v$ for that connect nodes $u$ and $v$ with probability $\frac{d_u d_v}{2M}$.

\begin{equation}
    \label{equ: modularity}
    {\mathcal Q}=\frac{1}{2M}\sum_{u v}\left[{a}_{u v}-\frac{d_{u}d_{v}}{2M}\right]\delta(u, v)
\end{equation}

We use the optimisation procedure as formulated by \citet{tsitsulin2020graph} which first relaxes the cluster assignment matrix to a one-hot assignment matrix so that it becomes $\bm{\hat{Y}}\in \{0,1\}^{N \times k}$. The optimal $\bm{\hat{Y}}$ that maximises $\mathcal{Q}$ is the top-k eigenvectors of the modularity matrix $\bm{B}$. The modularity matrix can be reformulated as $\bm{B}=\bm{A}-{\frac{\bm{d} \bm{d}^\top}{2M}}$ which allows us to decompose the computation $\mathrm{Tr}(\bm{\hat{Y}}^{\mathsf{T}}\bm{B \hat{Y}})$ as a sum of sparse matrix-matrix multiplication and rank-one degree normalization reduces the time complexity, so the computation becomes $\mathrm{Tr}(\bm{\hat{Y}}^{\top}\bm{A\hat{Y}} - \bm{\hat{Y}}^{\mathsf{T}}\bm{d}^{\top}\bm{d\hat{Y}})$. A collapse regularisation term is used where $||\bullet ||_{F}$ is the Frobenius norm. Therefore, the loss function for the model is given by Equation \eqref{equ: dmonloss}, noting that $N$ and $M$ are respectively defined as the number of nodes and edges in the graph $\mathcal{G}$.

\begin{eqnarray}
    \label{equ: dmonloss}
    & \mathcal{L}( \mathcal{G}; \bm{W}) = -\frac{1}{2M}\,\mathrm{Tr}(\bm{\hat{Y}}^{\mathsf{T}}\bm{A\hat{Y}} - \bm{\hat{Y}}^{\mathsf{T}}\bm{d}^{\mathsf{T}}\bm{d\hat{Y}}) \\ & \nonumber + \lambda_{r}(\frac{\sqrt{k}}{N} \left| \left| \sum_{u}{\bm{\hat{Y}}}_{u}^{\top}\right|\right|_{F}-1)
\end{eqnarray}

\subsection{Federated Learning}

The weight averaging algorithm from \citet{mcmahan2017communication} is used to create the federated part of the algorithm, which averages all clients parameters at the server then distributes the result back to each client for local training. If the federation was acting over heterogeneous graphs then this would not be appropriate as the weights would correspond to different feature so the activation's would not be in alignment. In these experiments, the contribution of each client to the average is weighted by the number of local data samples as the partitioning is node wise. The cycle is defined the global aggregation in Equation \eqref{equ: fedave} after a preset number of local epochs where the parameters are updated at each round $r$ using $\bm{W}^{(r+1)} \leftarrow \bm{W}^{(r)} - \eta \nabla \mathcal{L}$, where $\eta$ is the learning rate.  This is not be appropriate for all scenarios discussed as clients may experience different flows of data which will affect how much the parameters update between each communication round. In that scenario, techniques would need to be developed to moderate each clients contribution and the redistribution. However, here redistribution occurs for all clients after the global aggregation. The parameters $\bm{W}_m$ are initialised from a Xavier uniform distribution and this term hereafter is used to refer to all weights in the forward pass including both the GNN $\bm{F}()$, and the linear layer $\xi()$ with all bias terms. 

\begin{equation}
    \label{equ: fedave}
    \bm{\bar{W}} \leftarrow \sum_{c=1}^{C} \frac{N_c}{N} \cdot \bm{W}_c
\end{equation}

\subsection{Experiment Details}

In all experiments, the algorithm variations of the federation, centralised model and the isolated models that do not collaborate are subject to the following constraints. The experiments are designed so that in one, only one client may have access to each node, compared to the relaxing of this constraint. Each set of models are trained for the equivalent of five local epochs for every communication round and train for a total of 250 rounds. The cluster size regularisation parameter is set constant in the experiment $\lambda_r=1.0$ and the number of clients is varied between two and fifteen. 

All models are optimised with the Adam optimiser \cite{kingma2014adam} using a learning rate of $\eta = 0.001$, with no weight decay. The GNN functions that underpin each model reduce the dimensionality of features to $64$. The training/validation/testing splits are created by partitioning the adjacency matrix into the proportions $0.7/0.1/0.2$. All graphs considered are undirected which is maintained in the splits. Macro-F1 is used to assess performance to ensure that the evaluation is biased by label distribution. The best weight parameters over training are taken at the point of highest F1 validation score and performance is determined by the test set. All experiments are evaluated over three seeds and the average performance is reported. All experiments are performed on a single 2080 Ti GPU (12GB) with 96GB of RAM, and a 16core Xeon CPU. The $W$ Randomness Coefficient is used to assess the reliability of the results, where a lower value indicates higher consistency across random seeds over the experiment \cite{ugle2023leeney}. 

\begin{figure*}[hbt]
    \centering
    \includegraphics[width=0.9\textwidth]{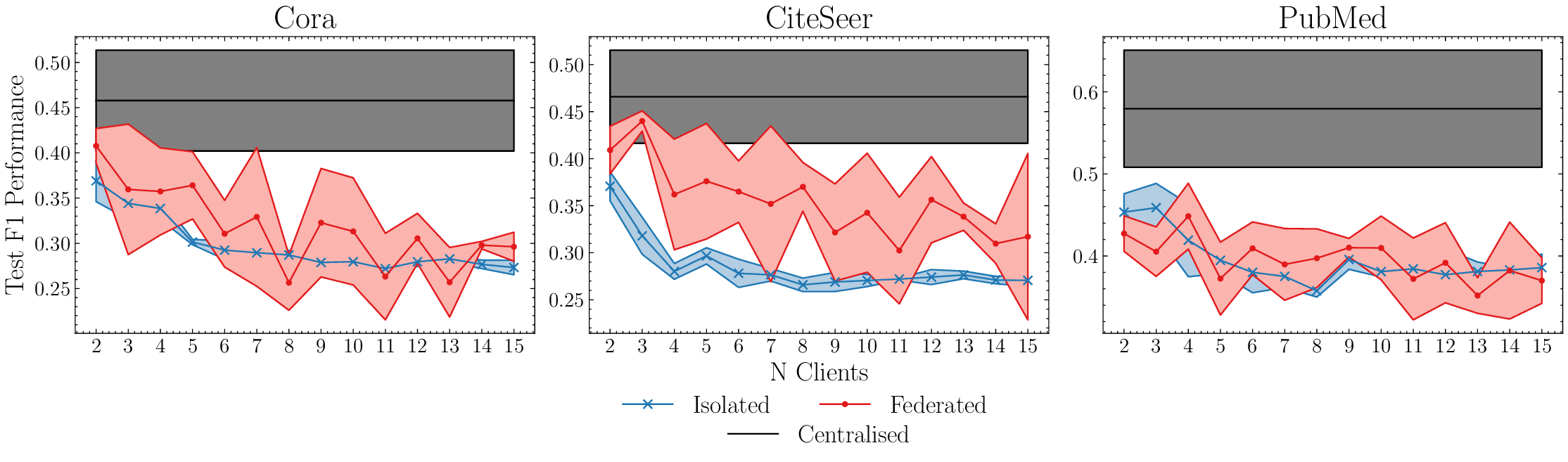}
    \caption{The performance comparison of the different algorithms at unsupervised graph clustering under a federated learning setup with distinct node partitions ($W$: 0.212).}
    \label{fig: exp1}
\end{figure*}

\begin{figure*}[hbt]
    \centering
    \includegraphics[width=0.9\textwidth]{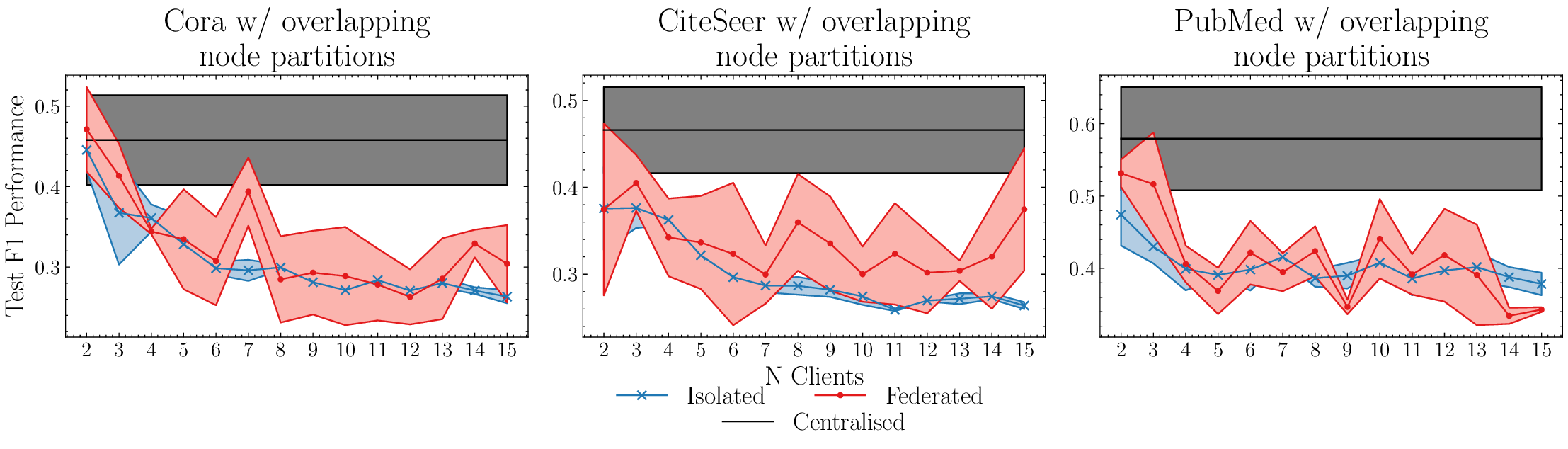}
    \caption{This shows performance comparison for the federated, isolated and centralised models where each client may own the same node ($W$: 0.259).}
    \label{fig: exp2}
\end{figure*}

\section{Discussion}

\subsection{Experiments}

The centralised model performs better than any other number of clients. The lighter region of colour in Figure \ref{fig: exp1} shows performance variation due to random seed. In general, the federation performs better than the isolated models and performance degrades as number of clients is increased. In contrast, in Figure \ref{fig: exp2}, performance has increased for both the federation of clients and those that are isolated, so that for two clients performance is almost on par with the centralised model. In addition, the difference between the two client models is less than the previous experiment, which shows that the lost connectivity between clients' private nodes affects learning. Whilst still behind the centralised model that accesses all data, the potential of federated graph learning has been demonstrated at to be better than not collaborating for unsupervised clustering. A low $W$ randomness coefficient for both experiments means that these experiments are consistent across randomness and can therefore be trusted.

\subsection{Open Challenges}

\subsubsection*{\textbf{Diverse Models}}

Figure \ref{fig: exp1} highlights that the centralised model performs better than both the Legion and the clients that do not collaborate. This establishes the framework for federated community detection, however, only one federated GNN model is considered. A comprehensive study should consider a diverse range of GNN models. The dataset partitioning means that the local GNN cannot extract information from high-order neighbours due to the lost connectivity hence the drop in performance compared to the centralised model. Additionally, it may not be practical to share details of the model and hence share via weight aggregation. Using a loss function to minimise distance in the embedded space similar to \citet{makhija2022architecture} may be the solution to this.

\subsubsection*{\textbf{Non-IID Data}}

The isolated models performance is proportional to the edges lost in the dataset partitioning scheme and the increase in the number of clients. A biased partition of the data, where each client owns the majority of nodes in one cluster, would increase the drop in performance compared to the centralised model \cite{ma2022state}. An extension of this work that may be suitable for this scenario is to scale the weight averaging in proportion to the similarity function based on cluster size distributions. Alternatively, it may be pertinent to share cluster centers and modulate weight averaging by computing the average distances between clients cluster centers.

\subsubsection*{\textbf{Heterogeneous Partitions}}

The constraint of all clients learning on the same graph is used in this work to demonstrate the potential of unsupervised federated graph learning, however there are many applications that deal with complex graphs. MuMin \cite{nielsen2022mumin} is a multi-modal graph dataset for research and development of misinformation detection which is traditionally unsupervised task due to the unknown origin behind real world fake news. A federation that learns what feature information is relevant globally by feature alignment similar to \citet{zhuang2021towards} is likely to be useful for aggregating between these different modalities.

\subsubsection*{\textbf{Malicious Attack Vulnerability}}

In these experiments it is assumed that all parties are acting honestly and aren't training on bad data to intentionally worsen the average of the federation. If it is not possible to verify the integrity of all clients then it will be important for solutions to be developed in awareness of this. A solution to this might be to generate random graphs to send to each participant, rank each client ability to cluster then scale the weight averaging in proportion to the trustworthiness score given by the W randomness coefficient. 

\subsubsection*{\textbf{Efficient Communication}}

Communication and memory consumption can be a key bottleneck when applying federated learning algorithms in practice \cite{liu2022federated}. This scheme for aggregating the weights of clients means that the entire network needs to be communicated to a server which comes with a significant memory overhead. Efficient communication protocols will be needed for real world implementations, which could be achieved by learning a sparse mask for minimising overhead similar to ideas proposed by \citet{baek2022personalized}. An alternative may be to learn the topology of clients with a GNN which would allow clients to share over networks that aren't fully connected.

\section{Final Remarks}

In this work, we propose a framework that combines federated graph learning with community detection and validate the feasibility of it, whilst demonstrating that lost connectivity with privately held data is significant. This framework extends to a diverse range of federations that are encompassed by different graph partitions, with each relevant to a practical application. Many open challenges in this paradigm that are revealed by the initial set of experiments performed herein and we discuss potential solutions to these.

\bibliography{aaai24}

\end{document}